%% file: pprai-2024-rico-tw.tex
\documentclass{pprai}

\pdfoutput=1
\usepackage[utf8]{inputenc}

\usepackage[T1]{fontenc}
\usepackage{graphicx}
\usepackage{amsthm}
\usepackage{hyperref}
\usepackage{txfonts}
\usepackage{url}
\usepackage{algpseudocode}
\usepackage[dvipsnames]{xcolor}
\usepackage{subcaption}
\usepackage[export]{adjustbox}
\usepackage{pdfpages}

\title{Rico: extended TIAGo robot towards up-to-date social and assistive robot usage scenarios}
\headtitle{Rico: extended TIAGo robot ...}


\author{Tomasz Winiarski$^{[0000-0002-9316-3284]}$,\\ Wojciech Dudek$^{[0000-0001-5326-1034]}$, \\ Daniel Giełdowski$^{[0000-0002-4348-2981]}$}
\headauthor{T. Winiarski, W. Dudek, D. Giełdowski}
\affiliation{%
	Warsaw University of Technology\\
	Faculty of Electronics and Information Technology\\
	Institute of Control and Computation Engineering\\
	Nowowiejska 15/19, 00-665 Warsaw, Poland\\
	tomasz.winiarski@pw.edu.pl}


\begin{document}
	
	\includepdf{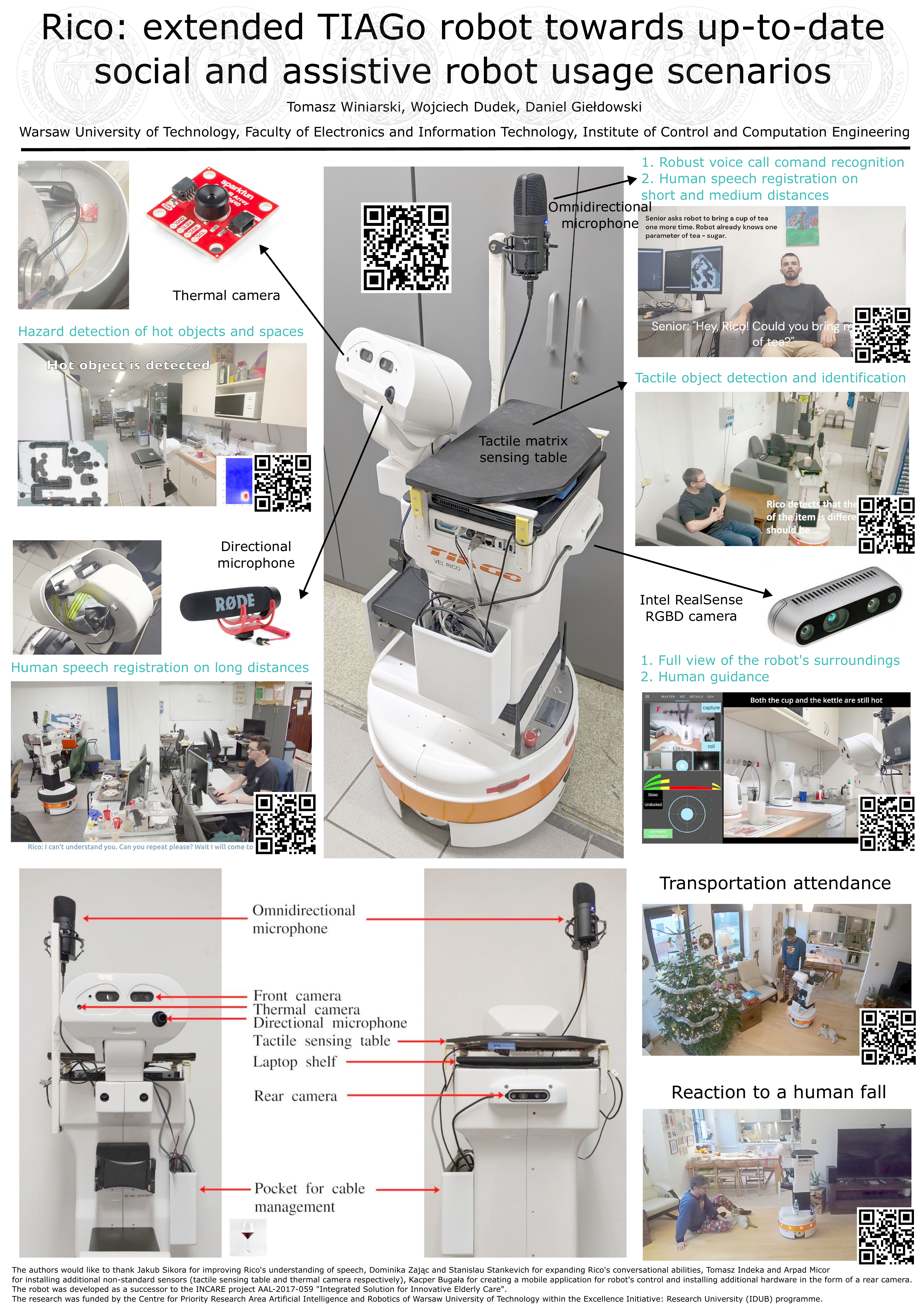}

	\maketitle
	
	\begin{abstract}
        Social and assistive robotics have vastly increased in popularity in recent years. Due to the wide range of usage, robots executing such tasks must be highly reliable and possess enough functions to satisfy multiple scenarios. This article describes a~mobile, artificial intelligence-driven, robotic platform Rico. Its prior usage in similar scenarios, the number of its capabilities, and the experiments it presented should qualify it as a~proper arm-less platform for social and assistive circumstances.
	\end{abstract}

\section{Introduction}
\label{sec:introduction}

    The ongoing demographic shifts require the development of novel technologies to assist a~significant number of elderly or disabled individuals. This demand is met using assistive and social robots \cite{leite2013social}. The development of artificial intelligence techniques, particularly efficient audio processing in the computational cloud \cite{recchuto2020cloud}, enables the successful implementation of a~voice communication interface as the primary HMI for this class of robots.
    
    When testing the performance of an assistive robot, it is important to consider relevant scenarios. During AAL Incare project\footnote{\url{http://aal-incare.eu/}}, Integrated Solution for Innovative Elderly Care, we have tested two potential tasks: assisting with item transportation\footnote{\url{https://vimeo.com/670252925}} and reaction to a~human fall\footnote{\url{https://vimeo.com/670246589}}. For this purpose, we used an early version of the PAL TIAGo-based Rico robot \cite{winiarski-intent-20}. The robot was also used for experiments with human-aware navigation \cite{karwowski-access-2023}. These experiments led to the discovery of potential problems arising from the current platform design.
    
    The above experiments resulted in improvements to Rico's robot design and software, briefly described in the following article. We improved the robot's communication capabilities by an additional microphone, implementing active listening behaviour \cite{jsikora-msc-21-eng} and using artificial intelligence for speech recognition, analysis and synthesis \cite{sstankevich-bsc-23-eng}. The addition of a~thermal imaging camera \cite{amicor-bsc-23-eng} made it possible to detect dangerously hot objects. The tactile sensing table \cite{tindeka-msc-24-eng} allows automatic detection and analysis of objects placed on the robot. Finally, the platform's manual control capabilities have been improved using an additional camera and a~smartphone-based GUI \cite{kbugala-bsc-23-eng}.
    
    The article starts with a~description of the Rico robot in its current state (sec.~\ref{sec:rico}). This robot (or parts thereof) was used in experiments described in sec.~\ref{sec:experiments}. The paper is finalised with conclusions (sec.~\ref{sec:conclusions}).

\section{Rico robot}
\label{sec:rico}

Rico is an extended version of a~TIAGo robot \cite{pages2016tiago}. It is a~ROS-based mobile robot used in multiple research projects. Its software includes basic ROS packages developed by PAL for the TIAGo robot extended by our platform-specific libraries. Therefore, we decided to employ the MeROS metamodel \cite{winiarski2023meros}\footnote{\url{https://github.com/twiniars/MeROS}} for its specification. Depending on the situation, tasks commissioned for social robots may require appropriate scheduling, including interrupting, postponing and resuming. This led to using the TaskER framework \cite{tasker2020} to control the robot's behaviour by switching between tasks dependent on their priority. Due to the possibility of damage to the robot, testing its capabilities is preceded by experiments performed in a~simulation environment, making it a~simulation-physical system. This means that the same software must be able to run on both the real and simulated versions of the platform. In order to quantitatively evaluate this system, we used the SPSysML \cite{dudek2023spsysml} philosophy.

\begin{figure}[h]
        \centering
        \includegraphics[width=\linewidth]{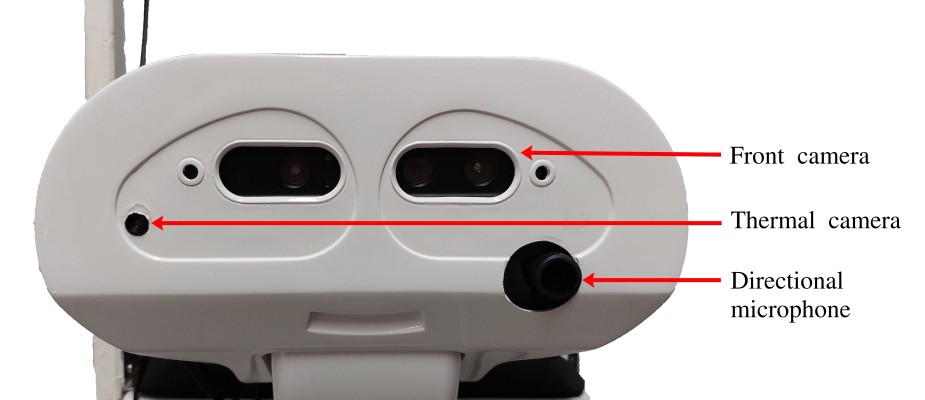}
        \caption{Hardware in Rico's head}
        \label{fig:rico_head_components}
    \end{figure}

    The previous version of the robot \cite{winiarski-intent-20} was extended by additional hardware. We managed to fit two extra sensors inside the robot's head - a~directional microphone and a~thermal camera (fig. \ref{fig:rico_head_components}). The reasons for such placement were keeping the integrity of the robot's design and the possibility of using them without turning the whole platform around, but only its head.

\begin{figure}[h]
        \begin{subfigure}[t]{0.24\linewidth}
            \centering
           \includegraphics[origin=c,width=\linewidth]{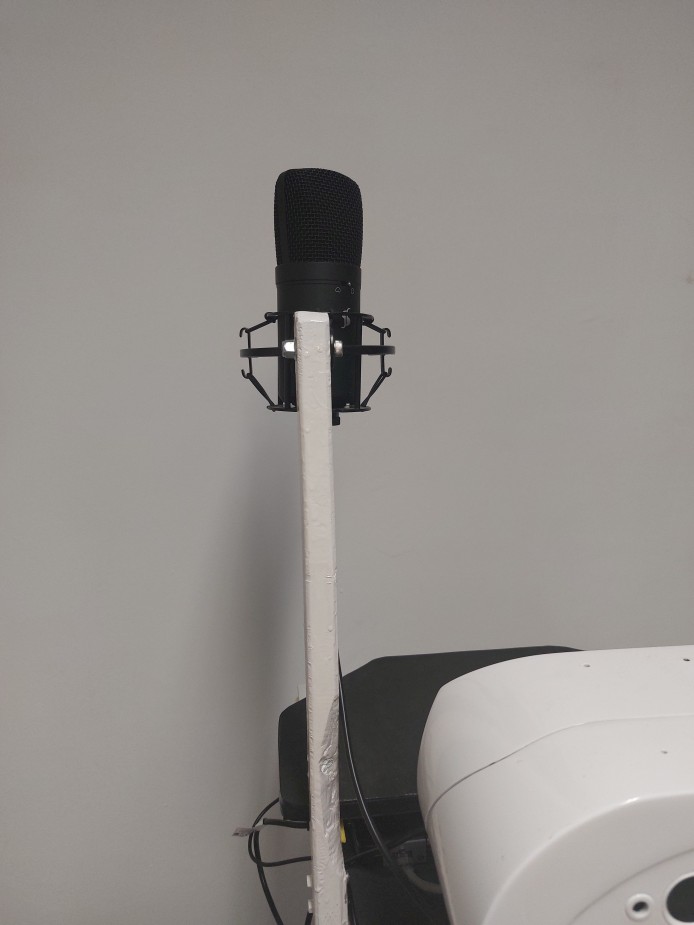}
            \caption{Omnidirectional microphone}
            \label{subfig:omni_microphone}  
        \end{subfigure}
        \begin{subfigure}[t]{0.24\linewidth}
            \centering
           \includegraphics[origin=c,width=\linewidth]{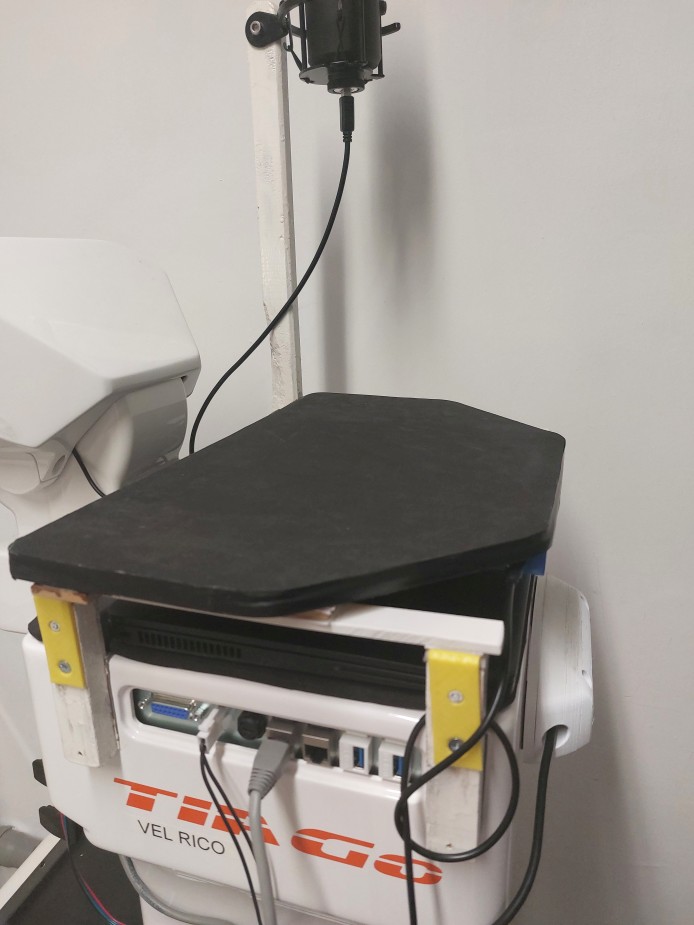}
            \caption{Tactile sensing\\table with laptop\\shelf below}
            \label{subfig:tactile_table}  
        \end{subfigure}
        \begin{subfigure}[t]{0.24\linewidth}
            \centering
           \includegraphics[origin=c,width=\linewidth]{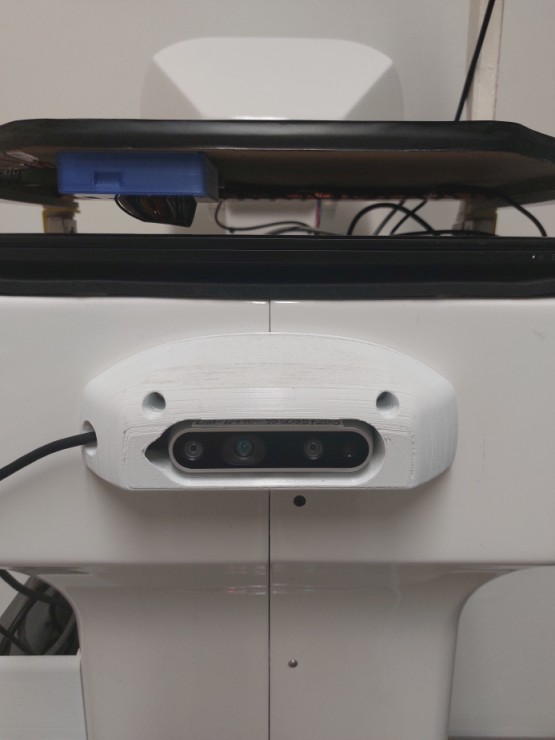}
            \caption{Rear RGBD\\camera}
            \label{subfig:back_camera}  
        \end{subfigure}
        \begin{subfigure}[t]{0.24\linewidth}
            \centering
           \includegraphics[origin=c,width=\linewidth]{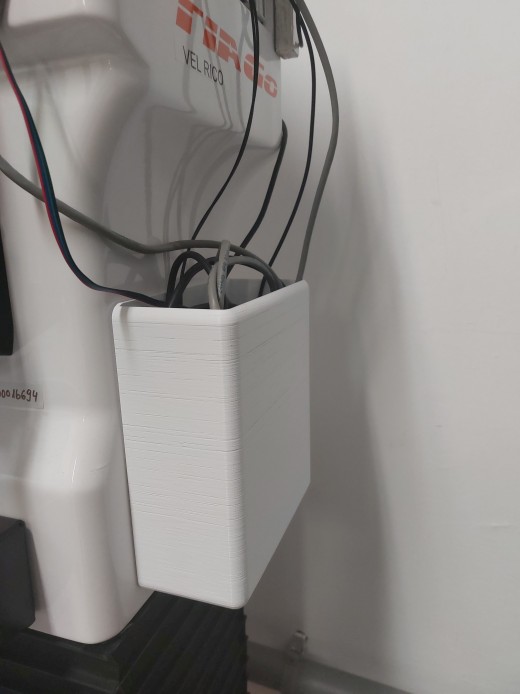}
            \caption{Pocket for cable management}
            \label{subfig:cable_pocket}  
        \end{subfigure}
        \caption{New hardware attached to Rico's body}
        \label{fig:rico_new_hardware}
    \end{figure}

\begin{figure}[h!]
        \centering
        \includegraphics[width=\linewidth]{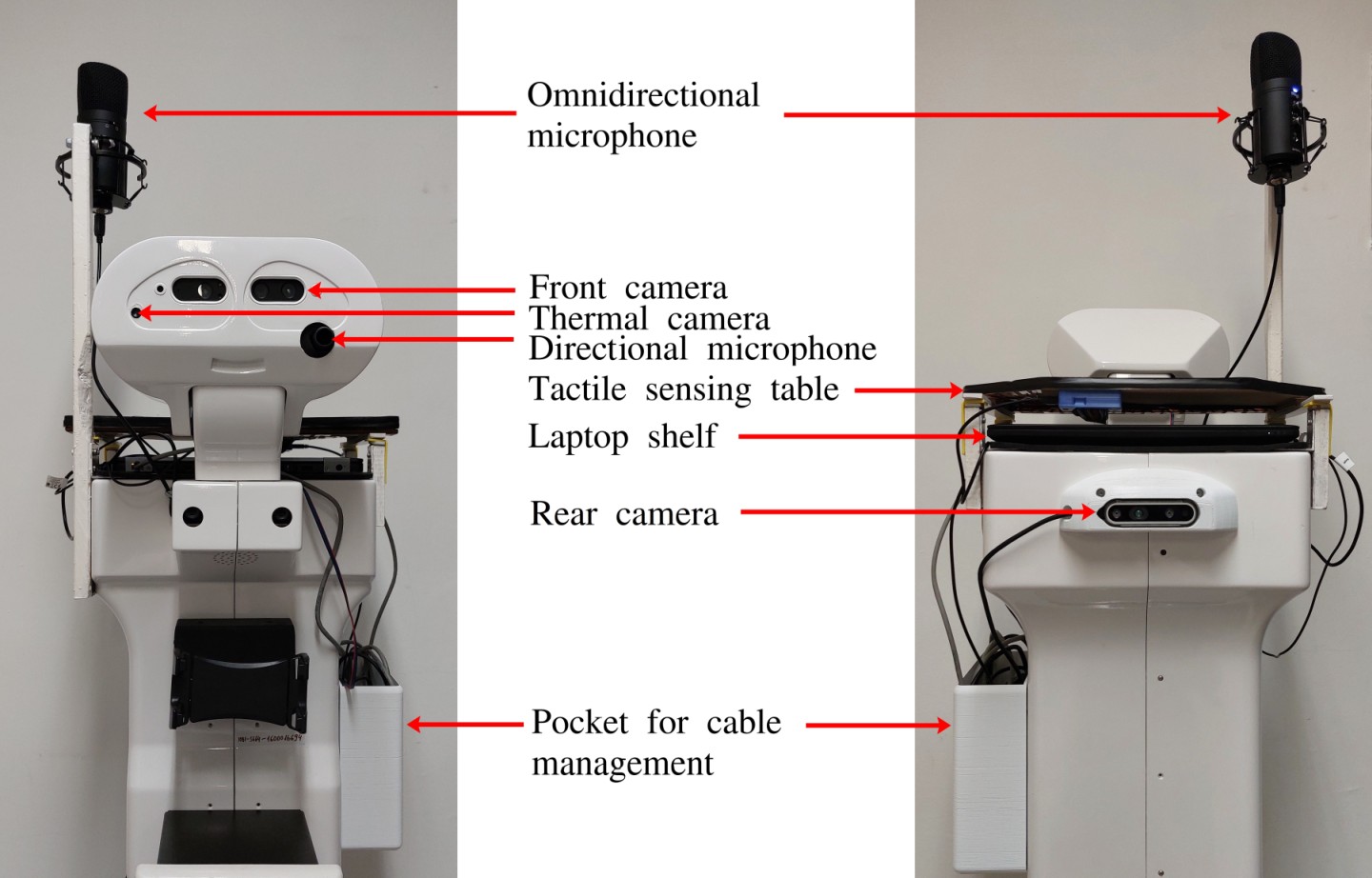}
        \caption{All of Rico's hardware}
        \label{fig:rico_components}
    \end{figure}

    In addition to the directional microphone, we kept the omnidirectional one mounted on the pole above the robot's head (fig. \ref{subfig:omni_microphone}). The tactile sensing table was placed on the robot's top, above the original surface (fig. \ref{subfig:tactile_table}). It includes more than 200 tactile tiles capable of detecting the pressure on their surface. We also left enough space under the table to accommodate the laptop used to process the data from some of the sensors. The final improvement included the addition of an RGBD realsense camera on the back of the robot (fig. \ref{subfig:back_camera}). This camera provides high-quality images taken from the opposite side than the camera in the head, which may be helpful, e.g., while manually controlling the robot (for example, through the mobile app). Due to the rising amount of wiring and additional controllers required for all of the new components we attached a~special pocket to the side of the robot (fig. \ref{subfig:cable_pocket}). The arrangement of Rico's hardware is presented in fig. \ref{fig:rico_components}.
    
\section{Experiments}
\label{sec:experiments}

    The first pool of experiments presented here was designed to address user's talk comprehension issues regardless of distance and orientation relative to the robot~\cite{jsikora-msc-21-eng}. By recording the user with both omnidirectional and directional microphone and processing both inputs simultaneously, we aimed to improve the robot's understanding of voice commands. By communicating with a~robot from different places, we were able to examine the quality of the recordings and verify the probability of understanding them. The result of the experiments was to get the robot to understand the command correctly more than 90\% of the time. Other behaviour was also developed in which the robot moves to the speaker in case of misunderstanding the command and faces them to better utilize the front microphone \footnote{\url{https://vimeo.com/603908325}}.

     Subsequent experiments were performed to improve the robot's conversational capabilities during task execution~\cite{sstankevich-bsc-23-eng}. This was done using the OpenAI GPT-3.5 model, which allows for greater flexibility in user commands compared to the previously used Google Dialogflow. In addition, it enabled the object transport scenario to be configured to learn new task parameters and actively use this knowledge in subsequent calls. For example, if a~question is asked about the amount of sugar when delivering tea, Rico will start asking for it every time it is asked for tea\footnote{\url{https://vimeo.com/863071575}}.

\begin{figure}[h!]
        \begin{subfigure}[b]{0.49\linewidth}
            \centering
           \includegraphics[width=\linewidth]{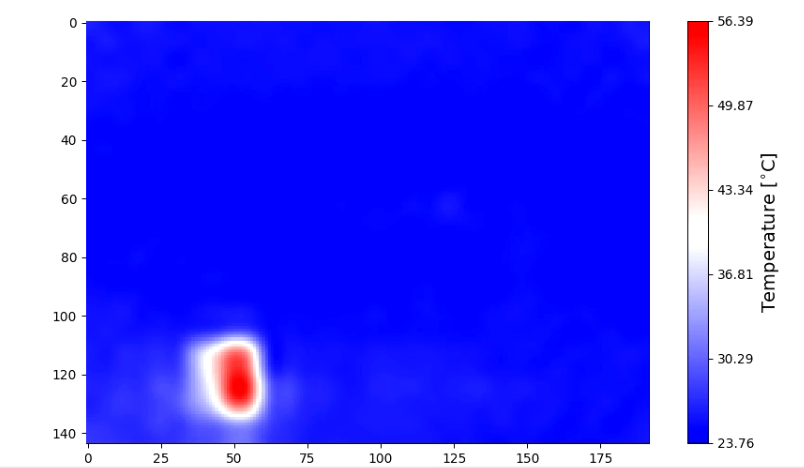}
            \caption{Detection result for cup with hot tea}
            \label{subfig:thermo_cup}  
        \end{subfigure}
        \begin{subfigure}[b]{0.49\linewidth}
            \centering
           \adjincludegraphics[width=\linewidth, trim={0 {.18\height} 0 {.18\height}}, clip]{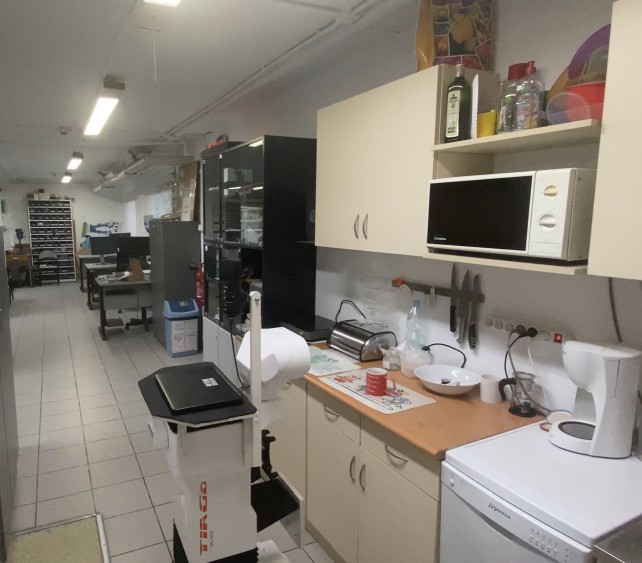}
            \caption{Robot position in environment}
            \label{subfig:thermo_person}   
        \end{subfigure}
        \caption{Heat detection using thermal camera}
        \label{fig:thermal_camera}
    \end{figure}

    Installing a~thermal imaging camera made it possible to detect ambient elements that stand out at higher temperatures\cite{amicor-bsc-23-eng}. During the experiments, detecting a~cup filled with hot liquid was undertaken (fig. \ref{subfig:thermo_cup}). The detection of human body temperature from various distances was also tested (fig. \ref{subfig:thermo_person}). A~new scenario was also implemented: the robot patrolled a~room by moving between two points. If a~hot object was detected, Rico would go to the base to communicate the hazard\footnote{\url{https://vimeo.com/836200374}}.

    To further refine the object transport scenario, experiments using a~tactile sensing table were performed \cite{tindeka-msc-24-eng}. Its readings were used to detect the object's type, position and weight. During the exemplary transport experiment, if Rico detected that the mug was placed too close to the edge or its contents did not match the expected weight, the robot reported the problem to the user present\footnote{\url{https://vimeo.com/903840750}}.

\begin{figure}[h]
            \centering
           \includegraphics[width=.9\linewidth]{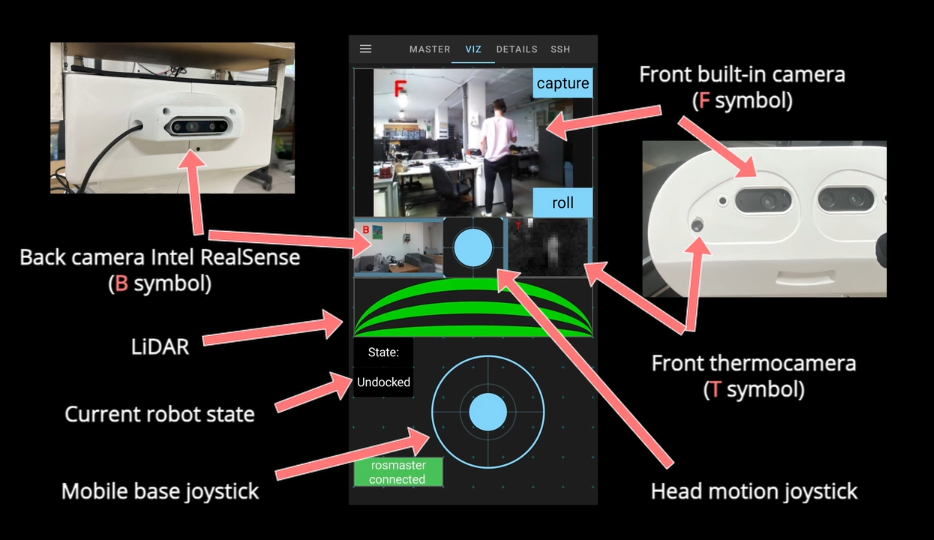}
        \caption{Mobile interface for robot teleoperation}
        \label{fig:rico_mobile}
    \end{figure}

   The last part of the experiments described here involved verifying a~smartphone GUI created to manually control the robot \cite{kbugala-bsc-23-eng}. Although not crucial for previously proposed scenarios, manual control is useful, e.g., when autonomous navigation is unavailable (for example, during the development of the robot's applications). The created interface is presented in fig. \ref{fig:rico_mobile}. During experiments, it was verified that the interface works properly, allowing for movement control of both the base and the head and for visualizing data from all of the robot's cameras and a~lidar sensor at satisfactory quality \footnote{\url{https://vimeo.com/865339765}}.

	\section{Conclusions}
	\label{sec:conclusions}
	
    Rico is a~mobile arm-less robot designed to carry out usage scenarios in social and assistive applications, thanks to many sensors implemented. It is currently under heavy development to assure reliability and extensive functionality. It uses multiple artificial intelligence algorithms, including a~language model, to improve the conversational system and multiple neural networks for detecting and recognising objects placed on the tactile sensing table. We are integrating every developed aspect of the robot's system to create a~final version with exhaustive capabilities.

	
	\section*{Acknowledgment}
	
	The authors would like to thank Jakub Sikora for improving Rico's understanding of speech, Dominika Zając and Stanislau Stankevich for expanding Rico's conversational abilities, Tomasz Indeka and Arpad Micor for installing additional non-standard sensors (tactile sensing table and thermal camera respectively), Kacper Bugała for creating a~mobile application for robot's control and installing additional hardware in the form of a~rear camera.

    The robot was developed as a~successor to the INCARE project AAL-2017-059 "Integrated Solution for Innovative Elderly Care". The research was funded by the Centre for Priority Research Area Artificial Intelligence and Robotics of Warsaw University of Technology within the Excellence Initiative: Research University (IDUB) programme.
	
	
	\bibliography{pprai-2024-rico-tw}
	\bibliographystyle{pprai}
	
\end{document}